\documentclass[sigconf]{acmart}

\usepackage[switch]{lineno}
\usepackage{booktabs}
\usepackage{subfigure}
\usepackage{amsthm}
\usepackage{multirow}

\theoremstyle{definition}

\AtBeginDocument{%
  \providecommand\BibTeX{{%
    \normalfont B\kern-0.5em{\scshape i\kern-0.25em b}\kern-0.8em\TeX}}}

\begin{document}

\title{Domain Adaption for Knowledge Tracing}


\author{Song~Cheng$^1$, Qi~Liu$^{1,*}$, Enhong~Chen$^1$
}
\affiliation{
	\institution{
		$^1$Anhui Province Key Laboratory of Big Data Analysis and Application, School of Computer Science and Technology, University of Science and Technology of China, \\ chsong@mail.ustc.edu.cn; \{qiliuql,cheneh\}@ustc.edu.cn;
	}
}

\renewcommand{\shortauthors}{Submitted for blind review.}


\begin{abstract}
	With the rapid development of online education system, knowledge tracing which aims at predicting students' knowledge state is becoming a critical and fundamental task in personalized education. Traditionally, existing methods are domain-specified. However, there are a larger number of domains (e.g., subjects, schools) in the real world and the lacking of data in some domains, how to utilize the knowledge and information in other domains to help train a knowledge tracing model for target domains is increasingly important. We refer to this problem as {\em domain adaptation for knowledge tracing} (DAKT) which contains two aspects: (1) how to achieve great knowledge tracing performance in each domain. (2) how to transfer good performed knowledge tracing model between domains. To this end, in this paper, we propose a novel adaptable framework, namely {\em adaptable knowledge tracing} (AKT) to address the DAKT problem. Specifically, for the first aspect, we incorporate the educational characteristics (e.g., slip, guess, question texts) based on the {\em deep knowledge tracing} (DKT) to obtain a good performed knowledge tracing model. For the second aspect, we propose and adopt three domain adaptation processes. First, we pre-train an auto-encoder to select useful source instances for target model training. Second, we minimize the domain-specific knowledge state distribution discrepancy under maximum mean discrepancy (MMD) measurement to achieve domain adaptation. Third, we adopt fine-tuning to deal with the problem that the output dimension of source and target domain are different to make the model suitable for target domains. Extensive experimental results on two private datasets and seven public datasets clearly prove the effectiveness of AKT for great knowledge tracing performance and its superior transferable ability.
\end{abstract}





\maketitle

\begin{figure}[t]
	\centering
	\includegraphics[scale=1.1]{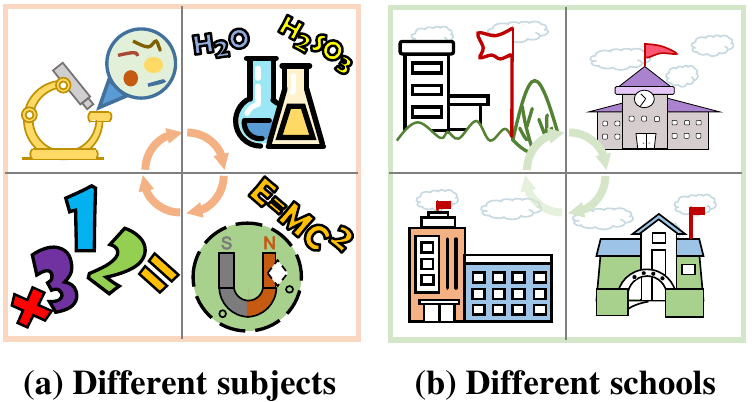}
	\vspace{-1em}
	\caption{A toy example of different domains.}
	\label{fig:domains}
\end{figure}

\section{Introduction}
In recent decades, a large number of computer-aided education systems (CAE) developed rapidly, such as massive open online courses~\cite{anderson2014engaging} and intelligent tutoring systems~\cite{burns2014intelligent}. They can give students an open access to the world-class instruction and a reduction in the growing cost of learning, which attracts lots of relevant users worldwide to join in these platforms. With the abundant learning records data of students collected from these platforms (e.g., Khan Academy, Edx), many researchers enrol to research and monitor students' learning process~\cite{chen2017tracking,Khajah2016HowDI}, to recommend adaptative learning materials and arrange personalized plans for them. Actually, it is helpful for students to learn the knowledge concepts which are not mastered, and to consolidate the knowledge concepts which are already mastered. To achieve this goal, the researchers introduce the {\em knowledge tracing} (KT) problem which tracking and estimating the knowledge state of the student, i.e., how much the student masters the knowledge concepts, to predict the score of him on the next question.

In the literature, there are many efforts in KT problem. Existing methods can be divided into two categories: traditional and deep approaches. The representative works of conventional knowledge tracing methods such as {\em bayesian knowledge tracing} (BKT)~\cite{corbett1994knowledge} and {\em performance factors analysis} (PFA)~\cite{pavlik2009performance} have been widely applied in many application scenarios. As for the deep approaches, {\em deep knowledge tracing} (DKT)~\cite{Piech2015DeepKT} is the first deep learning based method, which outperforms all of the conventional methods because of its nonlinear input-to-state and state-to-state transitions. Therefore, many variants (e.g., DKT+forgetting~\cite{nagatani2019augmenting}, DKT-tree~\cite{yang2018implicit}) of DKT have been proposed to improve KT performance.

Though great success has been achieved by BKT, DKT, etc., there still exist some issues that limit the application of them. First, as shown in Figure~\ref{fig:domains}, in the real world, there are many different subjects, schools and even grades that can be seen as different domains. However, existing KT models are domain (e.g., subject, schools, grades) specified methods, and the specific (e.g., math) model cannot be applied to other domains (e.g., physics) directly. Therefore, it is labor and resource-intensive to build so many models for all of the application scenarios. Second, the another issue is that there are some domains (e.g., some schools) usually lack abundant data for training models. Generally, how to utilize the knowledge and information in other domains to help train a KT model for target domains is increasingly important. We note to this problem as {\em domain adaptation for knowledge tracing} (DAKT) which contains two aspects: (1) how to achieve great knowledge tracing performance in each domain? (2) how to transfer good performed knowledge tracing model between domains?

\begin{figure}[t]
	\centering
	\includegraphics[scale=0.88]{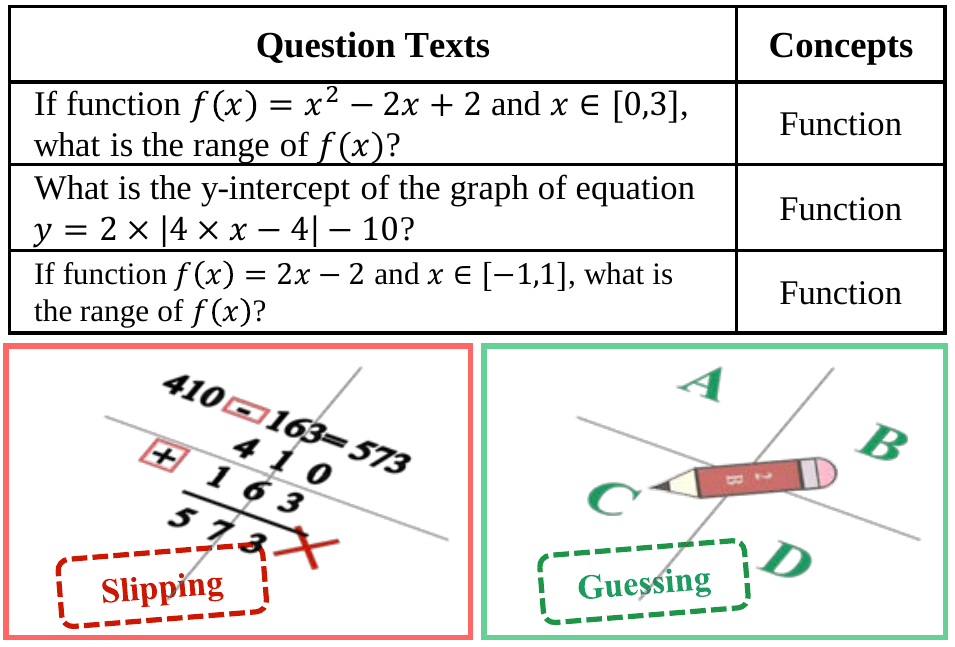}
	\vspace{-1.5em}
	\caption{Example of the question texts, slip and guess cases.}
	\vspace{-1.5em}
	\label{fig:toyexample}
\end{figure}

In this paper, to solve DAKT problem introduced above, we propose an {\em adaptable knowledge tracing} (AKT) framework. Specifically, on one hand, for achieving great knowledge tracing performance in each domain,  following previous works~\cite{cheng2019enhancing,Yeung2019DeepIRTMD}, we model some educational characteristics (e.g., slip, guess, question texts) which are important to model student learning process and ignored by existing methods. For example, as shown in Figure~\ref{fig:toyexample}, it is slipping that the student gets a wrong answer because he mistakes the minus sign for a plus sign. It is guessing that he chooses a correct option by rotating the pencil to get a random recommendation (e.g., {\em C}). It is better to embed question texts for representation than knowledge concepts since question texts are totally different though the knowledge concepts are the same. On the other hand, for achieving good information transferring between domains, we propose and adopt three domain adaptation processes. First, with reconstruction error, we pre-train an auto-encoder to select useful source instances for target model training. Second, we minimize the domain-specific knowledge state distribution discrepancy which is caused by the different distributions of student score $r_t$ and question texts $q_t$ under {\em maximum mean discrepancy} (MMD)~\cite{Borgwardt2006IntegratingSB} to achieve domain adaptation. Third, we adopt fine-tuning~\cite{yosinski2014transferable,kermany2018identifying} to deal with the problem that the output dimensions of source and target domains are different to make the model suitable for target domain. Extensive experimental results on two private datasets and seven public datasets clearly demonstrate the effectiveness of AKT for great knowledge tracing results and its superior transferable ability.

The main contributions in this work are summarized as follows:
\begin{itemize}
	\item[(1)]To our best knowledge, it is the first time that we propose the {\em domain adaptation for knowledge tracing} (DAKT) problem  for the wide application of knowledge tracing.
	\item[(2)]We model the educational characteristics (e.g., slipping, guessing, question texts) to achieve good knowledge tracing performance for the first aspect of DAKT.
	\item[(3)]We propose and adopt instance selection via pre-training, domain discrepancy minimizing and fine-tuning to achieve great domain adaptation for the second aspect of DAKT.
\end{itemize}

\section{Preliminaries}
In this section, we introduce the knowledge tracing task and domain adaptation techniques, which are the cornerstone of our adaptable knowledge tracing framework.
\subsection{Knowledge Tracing}
Knowledge tracing (KT) is a fundamental task in online education platforms, which aims at tracing the knowledge state of students with their historical interaction performances. It is usually formulated as a supervised time sequence learning problem.

In an intelligent education system,
suppose there are |M| students and |Q| questions. We record
the interaction process of a certain student as $\mathcal{I} =
\{\varsigma_1, \varsigma_2, ..., \varsigma_T\}, \varsigma_t = (q_t, r_t)$, where $q_t \in Q$ represents the question practiced by the student at interaction step $t$, and $r_t$ denotes the corresponding score. Generally, if the student answers the question $q_t$ rightly, $r_t$ equals to $1$, otherwise $r_t$ equals to $0$.
Also, each question $q$ contains its corresponding knowledge
concept $k$.
Without loss of generality, in this paper, we represent each student’s interaction sequence as $\mathcal{I} = \{(k_1, \varsigma_1), (k_2, \varsigma_2), ..., (k_T, \varsigma_T)\}$ for knowledge tracing task. More formally, knowledge tracing can be formulated as follows:

\begin{definition}[Knowledge Tracing]
	Given the question interaction sequence $\mathcal{I}$ of each student and the materials
	of each question from interaction step $1$ to $T$, the goal
	is two-fold: (1) tracking the change of his knowledge states
	and estimating how much he masters knowledge
	concepts from step $1$ to $T$; (2) using his current knowledge state of him to predict the score
	$r_{T+1}$ on the next candidate question $q_{T+1}$.
\end{definition}

%


\subsection{Domain Adaptation}
Domain adaptation (DA) is a particular case of {\em transfer learning} (TL) which utilizes labeled data in one or more relevant source domains to execute new tasks in target domains. Given a labeled source domain $\mathcal{D}_S = \{(x_i, y_i)\}_{i=1}^{n_S}$ and a target domain $\mathcal{D}_T = \mathcal{D}_{Tl}\cup\mathcal{D}_{Tu}$, where $\mathcal{D}_{Tl} = \{(x_i,y_i)\}_{i=1}^{n_{Tl}}$ is the limited labeled part and $\mathcal{D}_{Tu} = \{x_i\}_{i=1}^{n_{Tu}}$ is the abundant unlabeled part of target domain, we formally introduce the definition of domain adaptation as follows:
\begin{definition}[Domain Adaptation] Assume the feature space, label space and conditional probability distributions of source and target domains are the same, i.e., $\mathcal{X}_S=\mathcal{X}_T$, $\mathcal{Y}_S=\mathcal{Y}_T$ and $\mathcal{Q}_S(y_S|x_S) = \mathcal{Q}_T(y_T|x_T)$. Meanwhile,  the margin distributions of source and target domain are different, that is $\mathcal{P}_S(x_S) \neq \mathcal{P}_T(x_T)$. The goal of DA is to leverage $\mathcal{D}_S$ with the help of $\mathcal{D}_T$ to learn a task function $f:x_T\mapsto y_T$ for $\mathcal{D}_T$.
\end{definition}

\begin{figure*}[t]
	\centering
	\includegraphics[scale=0.8]{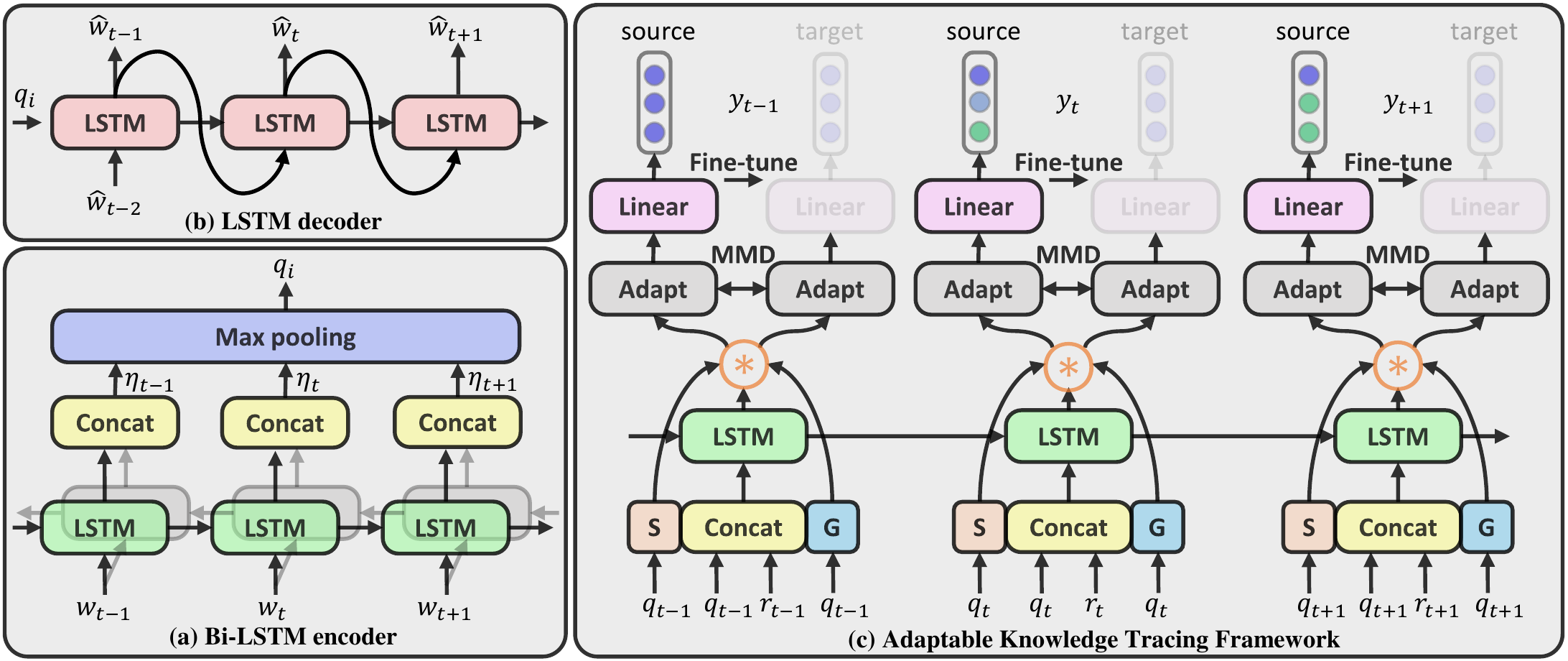}
	\caption{AKT model architecture.}
	\vspace{-1em}
	\label{fig:AKT}
\end{figure*}

\section{AKT: Modeling and Transferring}
In this section, we introduce AKT modeling and the transferring process which aim at solving DAKT problem in details. First, we give a formal definition of DAKT problem. Then, we describe the AKT architecture which introduces the educational characteristics for boosting KT task. Afterwards, we describe the transferring process of AKT, i.e., instance selection via pre-training, domain discrepancy minimizing and fine-tuning.

\subsection{DAKT Problem Definition}
Knowledge tracing is not the same as conventional machine learning tasks, since the input of KT task is the tuple $\varsigma_t = (q_t, r_t)$, where $\varsigma_t$ not only contains question $q_t$ but also students' score $r_t$. Meanwhile, $r_{t+1}$ is served as the label to be predicted next. This educational characteristic of KT makes it challenging to address {\em domain adaptation for knowledge tracing} (DAKT) problem. With different subjects, schools and even grades as domains, we give a formal definition of DAKT problem as follows:
\begin{definition}[DAKT Problem] Given the student' interaction sequence $\mathcal{I}_S$ in source domain and limited labeled part $\mathcal{I}_{Tl}$ of interaction sequence in target domain, the goal of DAKT is two-fold: (1) learning a good knowledge tracing model $\mathcal{M}$ by leveraging $\mathcal{I}_S$ in source domain $\mathcal{D}_S$; (2) transferring $\mathcal{M}$ to target domain $\mathcal{D}_T$ with the help of $\mathcal{I}_{Tl}$.
\end{definition}
In the following sections, we will address DAKT problem from two aspects: (1) how to introduce the educational characteristics (e.g., slip, guess, questions texts) for achieving better knowledge tracing; (2) how to conduct transferring processes to transfer good performed knowledge tracing model between domains for achieving domain adaptation.

\subsection{AKT Modeling}
As shown in Figure~\ref{fig:AKT} (c), the backbone of AKT model follows the architecture proposed by the original work of {\em deep knowledge tracing} (DKT) \cite{Piech2015DeepKT}. In the next, we will introduce the architecture of AKT framework.

\subsubsection{Question Texts Integrating} Since the question texts may be totally different though the knowledge concepts tested by them are the same, therefore, it will be better to represent the question by understanding its texts semantic than representing with knowledge concepts which is popular in previous works~\cite{Piech2015DeepKT, nagatani2019augmenting, yang2018implicit, Zhang2017Dynamic}. For integrating semantic of question texts, we adopt an unsupervised trained auto-encoder as shown in Figure~\ref{fig:AKT} (a) and Figure~\ref{fig:AKT} (b) to understand them, which is needed for transferring process described in next sections. The formal definition of the auto-encoder is as follows:
\begin{equation}
\begin{aligned}
&\text{encoder:}&q_i &= \pi_e(x_i),\\
&\text{decoder:}&\hat{x}_i &= \pi_d(q_i),
\end{aligned}
\label{eq:auto-enc-dec}
\end{equation} 
where $x_i = (w_1, w_2, ..., w_L)$ is the question text embedding sequence, $L$ is the length of the question text, and $\hat{x}_i = (\hat{w}_1, \hat{w}_2, ..., \hat{w}_L)$ is the reconstructed result of $x_i$.

Specifically, for the encoder $\pi_e$, its output $q_i$ can be considered as a higher-level semantic and robust representation for $i$-th question. Note that the input of the encoder is the question text which is a word sequence. In this case, we choose Bi-LSTM shown in Figure~\ref{fig:AKT} (a) as the encoder $\pi_e$ because it can make use of the most of contextual content information of question text from both forward and backward directions \cite{Huang2015BidirectionalLM}. Given $i$-th question text embedding sequence $x_i$, at each step $t$, the forward hidden state $\overrightarrow{h}_t^{enc}$ and backward hidden state $\overleftarrow{h}_t^{enc}$ are updated with current word $w_t$, and previous hidden state $\overrightarrow{h}_{t-1}^{enc}$ for forward direction or $\overleftarrow{h}_{t+1}^{enc}$ for backward direction in a recurrent formula as:
\begin{equation}
\begin{aligned}
&\overrightarrow{h}_t^{enc} = \mathrm{LSTM}(\overrightarrow{h}_{t-1}^{enc}, w_t; \overrightarrow{\theta}_{LSTM}^{enc}), \\
&\overleftarrow{h}_t^{enc} = \mathrm{LSTM}(\overleftarrow{h}_{t+1}^{enc}, w_t; \overleftarrow{\theta}_{LSTM}^{enc}),
\end{aligned}
\label{eq:bi-lstm}
\end{equation}
where recurrent formula follows Hochreiter et al. \cite{Hochreiter1997LongSM}.

As hidden state at each direction only contains one-side context, it is beneficial to concatenate the hidden state of both directions into one vector to capture the linguistic information at each step. Therefore, we model the semantic representation at $t$-th step as:
\begin{equation}
\eta_t = \mathrm{concatenate}(\overrightarrow{h}_{t}^{enc}, \overleftarrow{h}_{t}^{enc}),
\label{eq:u-concat}
\end{equation}
and then combine all representations at all time steps into one single vector $q_i$ with max-pooling, more formally:
\begin{equation}
q_{ij} = \mathrm{max}(\eta_{1j}, \eta_{2j}, ..., \eta_{Lj})
\label{eq:max}
\end{equation}

As for the decoder $\pi_d$, since the input $x_i$ that we aim to reconstruct is a word sequence, and LSTMs have yielded state-of-the-art results for many problems, such as image captioning \cite{Vinyals2014ShowAT}, natural language processing \cite{Sutskever2014SequenceTS}, and handwriting recognition \cite{Graves2009ANC}. Therefore, for the sake of simplicity, we adopt a single layer LSTM neural network as the decoder structures similar to sequence autoencoders \cite{Sutskever2014SequenceTS}. Comparatively, the input at each $t$-th time of the decoder $\pi_d$ is different from the general LSTM model, which is, $\pi_d$ leverages the $(t-1)$-th step output $\hat{w}_{t-1}$ as the $t$th step input. More formally, the definition of the $t$-th step hidden state can be formulated as:
\begin{equation}
\begin{aligned}
&h_t^{dec} = \mathrm{LSTM}(h_{t-1}^{dec}, \hat{w}_{t-1}; \theta_{LSTM}^{dec}),\\
&\hat{w}_t = \mathrm{sigmoid}(\mathrm{W}_{dec}\cdot h_t^{dec} + \mathrm{b}_{dec}),
\end{aligned}
\label{eq:dec-lstm}
\end{equation}
where $\theta_{LSTM}^{dec}$, $\mathrm{W}_{dec}$ and $\mathrm{b}_{dec}$ are the parameters of the decoder.

Without the loss of generality, we assume $r_t$ in each interaction $\varsigma_t = (q_t, r_t)$ is either 0 or 1. Methodology-wise, to be consistent with DKT when integrating semantic understanding of question texts~\cite{su2018exercise}, we first give a zero vector $\textbf{0} = (0, 0, ..., 0)^\mathrm{T}$ which has the same dimension $d_q$ as $q_t$, then we concatenate $q_t$ and $\textbf{0}$ to obtain the interaction representation $\varsigma_t$ at each time step $t$ as:
\begin{equation}
\varsigma_t = 
\begin{cases}
[q_t\oplus\textbf{0}],\quad \ \ & r_t = 1,\\ 
[\textbf{0}\oplus q_t], \quad \ \ & r_t = 0,
\end{cases}  
\label{eq:concatenate}
\end{equation}
where $\oplus$ is the operation of concatenating two vectors.


\subsubsection{Slipping and Guessing Modeling}To more precisely predict student's knowledge state and give strong interpretability for it, we introduce the slipping factor $s_t$ and the guessing factor $g_t$ factors of question $q_t$ to KT process. Different from conventional random sampling approach \cite{Wu2015CognitiveMF}, we propose to exploit the slipping and guessing factors from the semantic representation of the question texts. Specifically, as shown in Figure~\ref{fig:toyexample}, following previous work~\cite{Yeung2019DeepIRTMD,cheng2019enhancing}, we propose two single layer neural networks $S$ and $G$ as shown in Figure~\ref{fig:AKT} (c) to model the slipping and guessing factors as follows:
\begin{equation}
\begin{aligned}
&s_t = \mathrm{S}(q_t), \\
&g_t = \mathrm{G}(q_t),
\end{aligned}
\label{eq:slipguess}
\end{equation}
where $s_t$ and $g_t$ have the same dimension of $d_h$.

\subsubsection{Knowledge State Tracing}With the interaction sequence $\mathcal{I} = \{\varsigma_1, \varsigma_2, ..., \varsigma_{N}\}$ of a student where $N$ is the max sequence length. An LSTM architecture as described in DKT is then utilized to model the student learning process, and the hidden state $h_t\in\Bbb{R}^{d_h}$ at $t$-th step is updated as following formulas:
\begin{equation}
\begin{aligned}
&i_t = \sigma(\mathrm{W}_{\varsigma i}\cdot\varsigma_t+\mathrm{W}_{hi}\cdot h_{t-1}+\mathrm{b}_i),\\
&f_t = \sigma(\mathrm{W}_{\varsigma f}\cdot\varsigma_t+\mathrm{W}_{hf}\cdot h_{t-1}+\mathrm{b}_f),\\
&o_t = \sigma(\mathrm{W}_{\varsigma o}\cdot\varsigma_t+\mathrm{W}_{ho}\cdot h_{t-1}+\mathrm{b}_o),\\
&c_t = f_t c_{t-1} + i_t\tanh(\mathrm{W}_{\varsigma c}\cdot\varsigma_t + \mathrm{W}_{hc}\cdot h_{t-1} + \mathrm{b}_c),\\
&h_t = o_t\tanh(c_t),
\end{aligned}
\label{eq:LSTM}
\end{equation} 
where $i_\ast$, $f_\ast$, $c_\ast$, $o_\ast$ are the input gate, forget gate, memory cell and output gate of LSTM respectively, $\mathrm{W}_{\varsigma \ast}\in\Bbb{R}^{d_h\times 2d_q}$, $\mathrm{W}_{h\ast}\in\Bbb{R}^{d_h\times d_h}$ and $\mathrm{b}_{\ast}\in\Bbb{R}^{d_h}$ are the learned parameters.

Particularly, the matrix $\mathrm{W}_{\varsigma \ast}$ in Eq.~\ref{eq:LSTM} can be divided into a positive one $\mathrm{W}_{\varsigma \ast}^+\in\Bbb{R}^{d_h\times d_q}$ and a negative one $\mathrm{W}_{\varsigma \ast}^-\in\Bbb{R}^{d_h\times d_q}$~\cite{su2018exercise}, which can separately capture the influences of question $q_t$ with both right (e.g., $r_t=1$) and wrong (e.g., $r_t=0$) response. Therefore, it can naturally predict student knowledge states by integrating both question texts and responses.

After obtaining the representation $h_t$ of the interaction $\varsigma_t = (q_t, r_t)$, and the corresponding slipping $s_t$ and guessing $g_t$ factors of the question $q_t$, we simulate the student knowledge states $\kappa_t\in\Bbb{R}^{d_h}$ as follows: 
\begin{equation}
\kappa_t = (1-s_t)\circledast h_t + g_t\circledast(1-h_t),
\label{eq:state}
\end{equation}
where $\circledast$ is the element-wise multiply operation. $(1-s_t)\circledast h_t$ means the amount of knowledge state changing of a student when he masters the question and answers it successfully (i.e., without careless), while $g_t\circledast(1-h_t)$ represents the amount of knowledge state changing of the student when he guesses a right response without mastering \cite{Wu2015CognitiveMF}. That is, these are the two ways for a student to give a correct response.

Following DKT, we adopt a linear prediction layer to predict the probabilities $y_t\in\Bbb{R}^{Q}$ of answering correctly for all questions, with the simulated knowledge state $\kappa_t$ of the student. Comparatively, we add an adaptation layer before the prediction layer to achieve domain adaptation which is described in details in next sections. More formally, we define the probabilities $y_t$ as follows:
\begin{equation}
\begin{aligned}
&\alpha_t = \mathrm{adaptation}(\kappa_t; \Theta_{adp}), \\
&y_t = \mathrm{sigmoid}(\mathrm{W}_{out}\cdot\alpha_t + \mathrm{b}_{out}),
\end{aligned}
\label{eq:probabilities}
\end{equation}
where $\Theta_{adp}$ represents the parameters of adaptation layer, $\alpha_t$ is the output of adaptation layer at $t$-th step, $\mathrm{W}_{out}$ and $\mathrm{b}_{out}$ are the parameters of the prediction layer.
\subsection{Domain Adaptation}\label{transfer}
To achieve domain adaptation of KT to solve DAKT problem for wide application of it, we need to conduct some transferring process from source domain to target domain. Therefore, We propose to adopt three approaches from different aspects to address it greatly. Specifically, we first select the questions from source domain that are similar and useful for the target domain with pre-training, which we termed instance selection. Then, we add an adaptation layer before prediction layer to reduce the distribution divergence between domains by minimizing domain discrepancy. Moreover, since the output dimension is different between domains, we adopt fine-tuning technique to fine-tune the prediction layer for target domain. The details of them are described in the following subsections.
\subsubsection{Instance Selection via Pre-training}
The intuition of our proposed approach is that in a real scenario, if the data in the source domain are similar and useful for target domain, there should be a pair of encoder $\pi_e$ and decoder $\pi_d$ models whose reconstruction errors on source and target domain data are both small. Therefore, following previous work~\cite{tan2017distant}, we propose an auto-encoder \cite{Bengio2007LearningDA} as shown in Figure~\ref{fig:AKT} (a) and Figure~\ref{fig:AKT} (b) to select questions from source domain via minimizing reconstruction errors on the selected source domain questions and all target domain questions. The reconstruction error of the auto-encoder can be defined as:
\begin{equation}
\mathcal{R}(\hat{x}_i, x_i) = \frac{1}{L}\sum_{t=1}^{L}\lVert \hat{w}_t - w_t\rVert_2^2,
\label{eq:rec-error}
\end{equation}

To select the useful questions from source domain for target domain, we propose to pre-train the pair of $(\pi_e, \pi_d)$ to minimize the reconstruction errors on both selected questions in source domain and all the questions in target domain simultaneously. The formal definition of the objective function is as follows:
\begin{footnotesize}
	\begin{equation}
	\mathcal{J}_1(\pi_e, \pi_d, \textbf{u}_S) = \frac{1}{n_S}\sum_{i=1}^{n_S}u_S^i\mathcal{R}( \hat{x}_S^i-x_S^i) + \frac{1}{n_T}\sum_{i=1}^{n_T}\mathcal{R}( \hat{x}_T^i-x_T^i)+
	\Gamma(\textbf{u}_S),
	\label{eq:reconstruction}
	\end{equation}
\end{footnotesize}
where $x_S^i$ and $x_T^i$ are the questions in source and target domains, $\hat{x}_S^i, \hat{x}_T^i$ are the reconstructions of $x_S^i$ and $x_T^i$, $\textbf{u}_S = (u_S^1, u_S^2, ..., u_S^{n_S})$, and $u_S^i\in\{0,1\}$ is the selection indicators (e.g., 1 for selected, 0 for unselected) for the $i$-th question in source domain. The $\Gamma(\textbf{u}_S)$ in the objective function is a regulation term on $\textbf{u}_S$ to avoid the case that all values in $\textbf{u}_S$ is zero, it is defined as follows:
\begin{equation}
\Gamma(\textbf{u}_S) = -\frac{\lambda}{n_S}\sum_{i=1}^{n_S}u_S^i,
\label{eq:norm}
\end{equation}
where $\lambda \in [0, 1]$ controls the importance of the regularization term. It is obvious that more useful questions are selected, more robust semantic understanding can be learned by the pre-trained auto-encoder architecture.

To select useful questions in source domain, we update the parameters $\Theta_{auto}$ of auto-encoder and $\textbf{u}_S$ alternately. When $\textbf{u}_S$ is fixed, $\Theta_{auto}$ can be updated with back-propagation algorithm, comparatively, when $\Theta_{auto}$ is fixed, the solution of $\textbf{u}_S$ can be obtained as follows:
\begin{equation}
u_S^i = 
\begin{cases}
1,\quad \ \ & \mathcal{R}( \hat{x}_S^i-x_S^i) < \lambda,\\ 
0, \quad \ \ & \mathrm{otherwise},
\end{cases}  
\label{eq:selection}
\end{equation}

The reasonable explanation of this alternate training strategy is two folds: (1) fixing $\Theta_{auto}$ to update $\textbf{u}_S$, the useless questions in the source domain are removed; (2) updating $\Theta_{auto}$ with $\textbf{u}_S$ fixed, the auto-encoder is only trained on the selected questions.

\subsubsection{Domain Discrepancy Minimizing}
After obtaining the selected useful and similar questions in source domain, we can perform transfer learning across different domains. However, a model trained only on source domain usually leads to negative transfer, which will reduce the performance of the framework on target domain. This scenario is caused by the discrepancy of the probability distribution between the source and target domains data. Specifically, the margin distribution $\mathcal{P}_S(x_S)$ and $\mathcal{P}_T(x_T)$ of the question texts of source and target domains are different, moreover, the distribution of the score $r_t$ which mirrors the latent trait of the student is not the same too. That is, the student's hidden knowledge states that learned with $(q_t, r_t)$ are mismatched between two domains. As shown in figure~\ref{fig:AKT} (c), our intuition in this paper is to add an adaptation layer before output layer, to learn a sharing representation that minimizes the discrepancy between source knowledge state distribution $\mathcal{P}(\kappa_S)$ and target knowledge state distribution $\mathcal{P}(\kappa_T)$.

We consider adopting the widely applied distribution distance metric, {\em maximum mean discrepancy} (MMD)~\cite{Borgwardt2006IntegratingSB}. It is an effective criterion that compares distributions without their density functions. More formally, with two probability distributions $\mathcal{P}$ and $\mathcal{Q}$ on common feature space $\mathcal{X}$, MMD is defined as follows:
\begin{equation}
\mathcal{MMD}(\omega, \mathcal{P}, \mathcal{Q}) = \mathrm{sup}_{f\in\omega}(\Bbb{E}_{\tau\sim\mathcal{P}}[f(\tau)] - \Bbb{E}_{\upsilon\sim\mathcal{Q}}[f(\upsilon)]),
\end{equation}
where $\omega$ is a set of functions $f:\mathcal{X}\mapsto\Bbb{R}$, and $\Bbb{E}[\cdot]$ denotes the mean of the samples. Thus, based on the data $\textbf{x}_S = \{x_S^i\}_{i=1}^{n_S}$ and $\textbf{x}_T = \{x_T^j\}_{j=1}^{n_T}$ of source and target domains, MMD can be rewritten as follows:
\begin{equation}
\mathcal{MMD}(\textbf{x}_S, \textbf{x}_T)=\left\|\frac{1}{n_S}\sum_{i=1}^{n_S}\phi(x_S^i)-\frac{1}{n_T}\sum_{j = 1}^{n_T}\phi(x_T^j)\right\|_\mathcal{H},
\end{equation}
where $\phi(\cdot):\mathcal{X}\mapsto\mathcal{H}$ is referred as the feature space map, which maps the variable that in original feature space $\mathcal{X}$ to the {\em reproducing kernel hilbert space} (RKHS) $\mathcal{H}$ \cite{Borgwardt2006IntegratingSB}.

Moreover, our goal is not only to minimize the discrepancy between domains, but also to learn a robust knowledge tracing model. Therefore, the general form of the objective function can be expressed as:
\begin{equation}
\mathcal{J}_{AKT} = \mathcal{J}_{KT} + \gamma\mathcal{MMD}^2(\textbf{x}_S, \textbf{x}_T),
\end{equation}
where $\gamma \in [0, 1]$ is the regularization parameter that controls the importance of MMD, the $\mathcal{J}_{KT}$ denotes the loss of the knowledge tracing task \cite{Piech2015DeepKT}. More formally, it is defined as follows:
\begin{equation}
\mathcal{J}_{KT} = \sum_{i=1}^{n}\sum_{t}l(y^{(i)\mathrm{T}}\cdot\delta(x^{(i)}_{t+1}), r^{(i)}_{t+1}),
\end{equation}
where $n$ is the number of students, $l$ is the cross-entropy loss, $\delta(\cdot)$ is the one-hot encoding function which indicates $x_i$ is the $i$-th question.

Similarly, we separate $\mathcal{J}_{KT}$ and $\mathcal{MMD}^2(\cdot,\cdot)$, and minimize them alternately in two steps: (1) we first minimize $\mathcal{J}_{KT}$ with mini-batched stochastic gradient descent via back-propagation algorithm; (2) then $\mathcal{MMD}^2(\cdot,\cdot)$ is minimized with full-batched gradient descent. 

\subsubsection{Fine-tuning}
After training, AKT should be able to minimize the discrepancy between different domains and transfer knowledge from source domain to target domain. However, there is an issue which needs to be addressed, otherwise, we cannot apply our model trained on source data to target domain. Specifically, the number of the questions in the domains are different from each other, for example, there are totally $Q_S$ questions in source domain and $Q_T$ questions in target domain, so the model trained on source data would have a $Q_S$ dimension output layer, which mismatches the number $Q_T$ of target domain.

As we can see, the amounts of the questions of different domains are not the same. To transfer our AKT framework from source domain to target domain, only the output layer is required to be replaced and retrained for the specific target domain. It is termed as fine-tuning, which leads to effective knowledge transferring between different domains.

In summary, AKT has the following advantages. First, it provides a simple but effective way to capture the slipping and guessing factors of the questions, and leverage the semantic understanding of the texts to represent the question better for knowledge tracing performance improving. Second, it can achieve domain adaptation between source domain and target domain via pre-training, discrepancy minimizing and fine-tuning techniques.

\begin{figure}
	\centering  
	\subfigure[question text length of zx.math]{
		\label{fig:knowledge-dis}
		\includegraphics[scale=0.37]{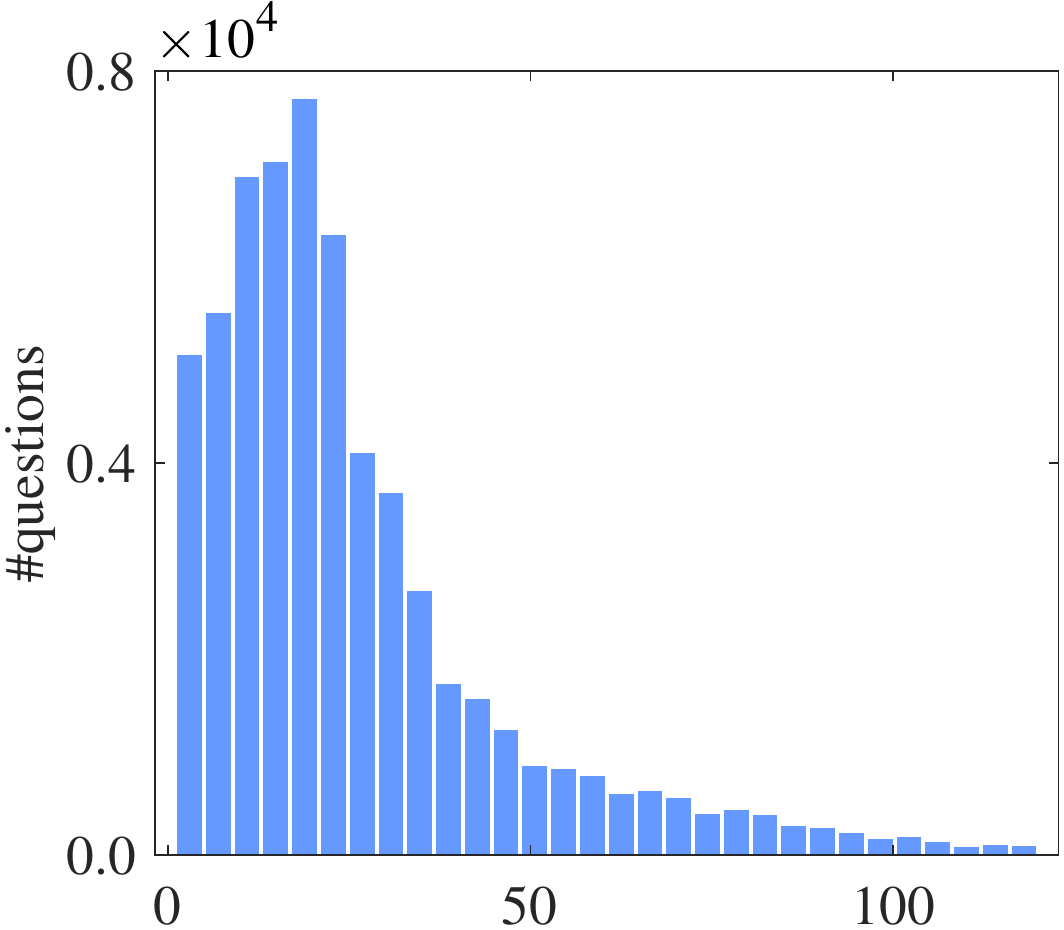}}
	\subfigure[question text length of zx.physics]{
		\label{fig:word-dis}
		\includegraphics[scale=0.37]{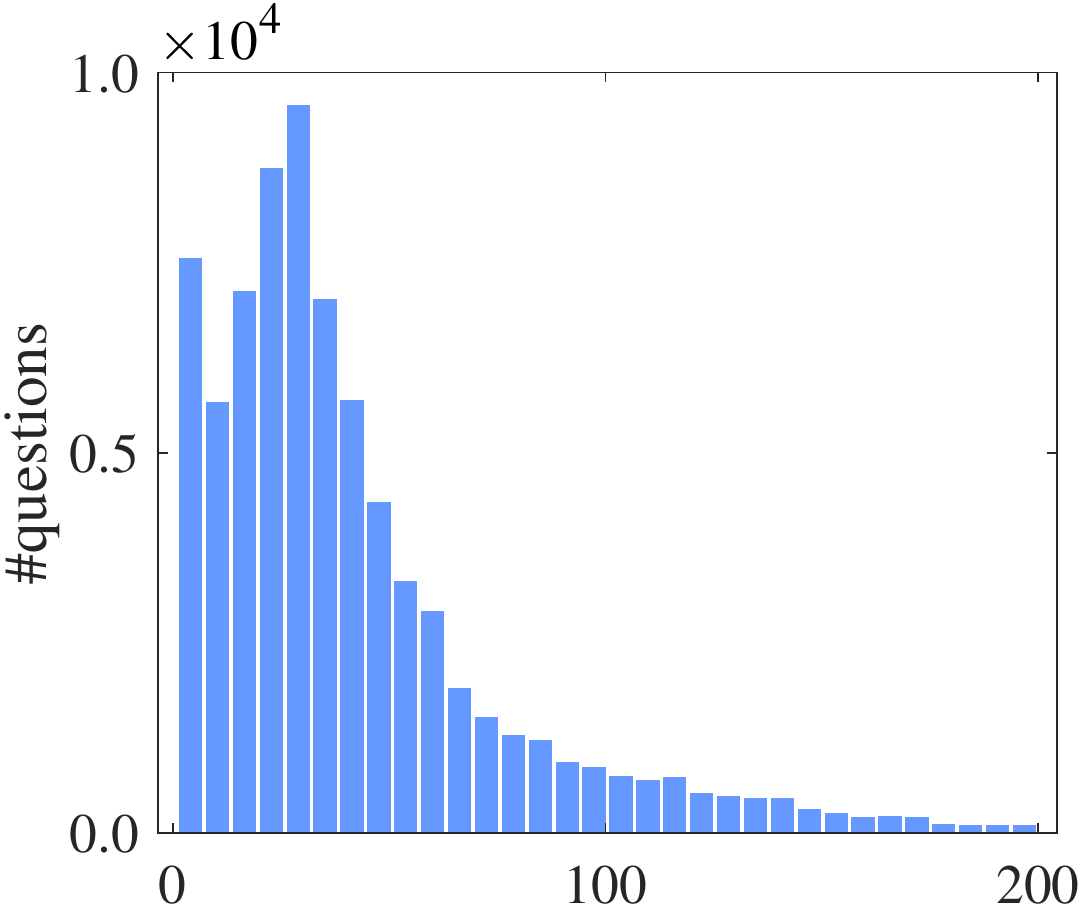}}
	\caption{Distribution of question text lengths.}
	\label{fig:distribution}
\end{figure}

\section{Experiments}
In this section, extensive experiments are conducted from two aspects to demonstrate the effectiveness of our AKT framework. First, the comparative experiments are conducted on two private datasets and seven public datasets to validate that AKT can model the educational characteristics (e.g., slipping, guessing, question texts) in education area well for boosting the knowledge tracing performance. Afterwards, we conduct transferring experiments on AKT with two subjects and four schools as different domains construct from the two private datasets, to demonstrate its superior transferable ability.

\subsection{Experimental Setup}
\subsubsection{Datasets}\label{datasets}The datasets used in this paper include two private datasets and seven public well-established datasets in KT task. The basic statistical information for all the datasets are summarized in Table~\ref{tab:statistics}. The detail descriptions are as follows:
\begin{itemize}
	\item[-]{\em zx.math.}
	This is a math subject dataset which is private and composed of the mathematical data collected from a number of senior high school in China. it is really a huge-scale dataset, we randomly selected 329,533 interactions to from 5,000 student with 60,752 questions.
	\item[-]{\em zx.physics.}
	This is a physics subject dataset which is private the same as {\em zx.math}, and collected from a number of senior high school in China. it is a huge-scale dataset too. we randomly selected 397,263 interactions to from 5,000 student with 75,141 questions.
	\item[-]{\em synthetic.}
	This dataset\footnote{https://github.com/chrispiech/DeepKnowledgeTracing/tree/master/data/synthetic} is simulated by piech et al. \cite{Piech2015DeepKT}, which simulated 2000 virtual students answering 50 questions both in training and testing dataset, which are drawn from five virtual concepts. In this dataset only, all students answer the same sequence of 50 questions.
	\item[-]{\em static2011.}
	This dataset\footnote{https://pslcdatashop.web.cmu.edu/DatasetInfo?datasetId=507} is from a college-level engineering statics course with 189,927 interactions from 333 student and 1,223 questions. In our experiments, we have adopted the processed data provided by Zhang et al. \cite{Zhang2017Dynamic}.
	\item[-]{\em kddcup2010.}
	This dataset\footnote{https://pslcdatashop.web.cmu.edu/KDDCup/downloads.jsp} is provided as the development dataset in KDD Cup 2010 competition, which comes from the Cognitive Algebra Tutor 2005-2006 Algebra system. After processing this dataset as Xiong et al. \cite{xiong2016going}, it contains 607,026 interactions from 574 student with 436 questions.
	\item[-]{\em junyi.}
	This dataset\footnote{https://datashop.web.cmu.edu/DatasetInfo?datasetId=1198} is collected from Junyi Academy\footnote{https://www.junyiacademy.org/}, which is an online education website. It is too large to effectively conduct experiments, so we randomly select 50,000 students with 7,868,952 interactions on 701 questions. It is still the largest dataset with a large number of interactions.
	\item[-]{\em ASSIST2009.}
	This dataset\footnote{https://sites.google.com/site/assistmentsdata/home/assistment-2009-2010-data/skill-builder-data-2009-2010} is collected from ASSISTments online tutoring system\footnote{https://www.assistments.org/}, which has been widely applied in KT tasks. Before conducting the experiments, we remove the duplicated records in the original dataset. The pruned dataset contains 325,673 interactions for 4,151 students with 110 questions.
	\item[-]{\em ASSIST2015.}
	This dataset\footnote{https://sites.google.com/site/assistmentsdata/home/2015-assistments-skill-builder-data} is collected from the same platform as ASSISTments2009, it contains 683,801 effective interactions from 19,840 students on 100 questions after processed.
	\item[-]{\em ASSIST2017.}
	This dataset\footnote{https://sites.google.com/view/assistmentsdatamining/dataset?authuser=0} is also collected from the same platform as ASSISTments2009, it is provided by the 2017 ASSISTments data mining competition, the average number of interactions per student is extremely high as it contains 686 students with 942,816 interactions on 102 questions.
\end{itemize}

\begin{table}[t]
	\centering
	\caption{The statistics of the dataset.}
	\begin{tabular}{l|lll}  
		\toprule\toprule
		\multirow{1}{*}{Dataset} & \multirow{1}{*}{\# Questions} & \multirow{1}{*}{\# Students} & \multirow{1}{*}{\# Interactions}\\
		\midrule\midrule
		zx.math			& 60,752 	 & 5,000 	 & 329,533		\\
		zx.physics 		& 75,141 	 & 5,000 	 & 397,263		\\
		Synthetic 		& 50 		 & 4,000	 & 200,000		\\
		Static2011 		& 1,223 	 & 333       & 189,297 		\\
		Kddcup2010 		& 436 		 & 574		 & 607,026		\\
		Junyi 			& 701		 & 50,000 	 & 7,868,952	\\
		ASSISTments2009 & 110		 & 4,151	 & 325,673		\\
		ASSISTments2015 & 100	 	 & 19,840	 & 683,801		\\
		ASSISTments2017 & 102 		 & 686 		 & 942,816		\\
		\bottomrule\bottomrule
	\end{tabular}
	\label{tab:statistics}
\end{table}

\begin{table*}[t]
	\centering
	\caption{Effectiveness of the slipping and guessing factors, the comparison of AUC of AKT-tx-tr with DKT on two private and seven public datasets. "s. dim" represents the hidden knowledge state dimension.}
	\label{table:slip_guess_AUC}
	\begin{tabular}{cc|ccccccccc}
		\toprule\toprule
		\multirow{1}{*}{Model}&\multirow{1}{*}{s.dim} & \multirow{1}{*}{zx.math} & \multirow{1}{*}{zx.physics} & \multirow{1}{*}{ASSIST2009} & \multirow{1}{*}{ASSIST2015} & \multirow{1}{*}{ASSIST2017} & \multirow{1}{*}{Synthetic} & \multirow{1}{*}{Static2011} & \multirow{1}{*}{Kddcup2010} & \multirow{1}{*}{Junyi} \\
		\midrule\midrule
		\multirow{4}{*}{DKT}
		& 10  &      0.7047 &      0.6984 &      0.8349 &      0.7155 &                  0.6534 &      0.8037 &      0.8186 &      0.7725 &      0.8783\\
		& 50  &      0.7248 &      0.7021 &      0.8369 & \bf{0.7314} &                  0.7029 & \bf{0.8251} &      0.8292 &      0.7989 &      0.8952 \\
		& 100 &      0.7255 &      0.6989 &      0.8398 &      0.7303 &                  0.6806 &      0.8171 &      0.8193 &      0.7724 &      0.8874\\
		& 200 &      0.7253 &      0.6993 &      0.8397 &      0.7301 &                  0.6861 &      0.8232 &      0.8217 &      0.7732 &      0.8901\\
		\cline{1-11}
		\multirow{4}{*}{AKT-tx-tr}
		& 10  &      0.7542 &      0.7416 &      0.8357 &      0.7075 &                  0.6534 &      0.7995 &      0.8429 &      0.7784 &      0.8864\\
		& 50  & \bf{0.7718} & \bf{0.7533} &      0.8440 &      0.7282 &             \bf{0.7181} &      0.8125 & \bf{0.8606} & \bf{0.8185} & \bf{0.9041}\\
		& 100 &      0.7623 &      0.7422 & \bf{0.8461} &      0.7245 &                  0.7056 &      0.8009 &      0.8488 &      0.8069 &      0.8957\\
		& 200 &      0.7621 &      0.7401 &      0.8447 &      0.7251 &                  0.7097 &      0.8179 &      0.8532 &      0.8086 &      0.8955\\
		
		\bottomrule\bottomrule
	\end{tabular}
\end{table*}

\begin{table*}[t]
	\centering
	\caption{Effectiveness of the slipping and guessing factors, the comparison of F1-score of AKT-tx-tr with DKT on two private and seven public datasets. "s. dim" represents the hidden knowledge state dimension.}
	\label{table:slip_guess_F1}
	\begin{tabular}{cc|ccccccccc}
		\toprule\toprule
		\multirow{1}{*}{Model}&\multirow{1}{*}{s.dim} & \multirow{1}{*}{zx.math} & \multirow{1}{*}{zx.physics} & \multirow{1}{*}{ASSIST2009} & \multirow{1}{*}{ASSIST2015} & \multirow{1}{*}{ASSIST2017} & \multirow{1}{*}{Synthetic} & \multirow{1}{*}{Static2011} & \multirow{1}{*}{Kddcup2010} & \multirow{1}{*}{Junyi} \\
		\midrule\midrule
		\multirow{4}{*}{DKT}
		& 10  &      0.7054 &      0.7025 &      0.8545 &      0.8506 &      \underline{0.3767} &      0.7878 &      0.8513 &      0.9031 &      0.8297\\
		& 50  &      0.7285 &      0.7163 &      0.8551 & \bf{0.8521} &      \underline{0.4687} &      0.7953 &      0.8661 &      0.9086 &      0.8405\\
		& 100 &      0.7338 &      0.7022 &      0.8563 &      0.8520 &      \underline{0.4163} &      0.7947 &      0.8548 &      0.9032 &      0.8341\\
		& 200 &      0.7314 &      0.7013 &      0.8564 &      0.8518 &      \underline{0.4216} & \bf{0.7990} &      0.8591 &      0.9039 &      0.8388\\
		\cline{1-11}
		\multirow{4}{*}{AKT-tx-tr}
		& 10  &      0.7793 &      0.7539 &      0.8543 &      0.8497 &      \underline{0.3424} &      0.7872 &      0.8600 &      0.9059 &      0.8402\\
		& 50  & \bf{0.7837} & \bf{0.7635} &      0.8580 &      0.8511 & \underline{\bf{0.4887}} &      0.7872 & \bf{0.8740} & \bf{0.9117} & \bf{0.8503}\\
		& 100 &      0.7751 &      0.7541 &      0.8576 &      0.8505 &      \underline{0.4427} &      0.7885 &      0.8708 &      0.9101 &      0.8421\\
		& 200 &      0.7736 &      0.7524 & \bf{0.8585} &      0.8506 &      \underline{0.4636} &      0.7937 &      0.8724 &      0.9106 &      0.8484\\
		\bottomrule\bottomrule
	\end{tabular}
\end{table*}

\subsubsection{AKT Setup}
For embedding, we incorporate {\em Word2vec}~\cite{Mikolov2013EfficientEO} on the whole corpus to get an initial word to vector mapping with size 100. The hidden state of the Bi-LSTM encoder is set to 50, to keep the output size of Bi-LSTM the same as the input size of the LSTM decoder. We adopt different regularization term importance via set the value of $\lambda$ as 0.0, 0.25, 0.5, 0.75, 1.0. We set the output size of slipping and guessing network to 100 which is the same as the hidden size of LSTM in AKT framework. To determine what dimension that our adaptation layer should have, we set the width of it as 64, 128, 256, 512, step by a power of two each time. The value of the penalty parameter $\gamma$ is set as 0.0, 0.25, 0.5, 0.75, 1.0.

Before any training step, we set the learning rate and batch size as $1\mathrm{e}{-5}$ and $64$ respectively. The parameters are randomly initialized with uniform distribution in the range between $-\sqrt{6/(nin/nout)}$ and $\sqrt{6/(nin/nout)}$ which follows Orr et at.~\cite{orr2003neural}, where $nin$ and $nout$ are the numbers of input and output size. Then at the training process, parameters are updated by Adam optimization algorithm~\cite{Kingma2014AdamAM}.

\subsubsection{Evaluation}
For the non-simulated data, we evaluate our results using 5-fold cross validation and in all cases, hyper-parameters are learned on training data. To evaluate the performance of the KT methods, the precise knowledge state can predict students' performance accurately. Thus, we use the area under the receiver operating characteristic (ROC) curve which is termed as AUC~\cite{Ling2003AUCAS}, and the F1 measure~\cite{Derczynski2016ComplementarityFA} which is the weighted harmonic mean of precision and recall. For both AUC and F1, the value 0.5 means the prediction results are guessed randomly, and the larger, the better. 

Note, we tune the parameters AKT and other models used in the next sections to achieve the best performance for the purpose of a fair comparison, and implement it with PyTorch on a Linux server with a Tesla K20m GPU.

\subsection{Educational Characteristics Modeling}

\subsubsection{Effectiveness of Slipping and Guessing}
To investigate the effectiveness of our framework for modeling the slipping and guessing factors, we conduct extensive experiments on AKT-tx-tr under AUC and F1 evaluation metrics, where "-" is a minus sign, which means AKT abandons the question texts and transferring process. For comparison, we adopt DKT as the baseline since it is similar to the backbone of AKT-tx-tr, that is both of them are based on LSTM neural network except the slipping and guessing modeling parts. 

The experimental results which clearly investigate the differentiation of the performance between AKT-tx-tr and DKT are shown in Table~\ref{table:slip_guess_AUC} and Table ~\ref{table:slip_guess_F1}. Specifically, we compare DKT with AKT-tx-tr under different hidden layer dimensions (e.g., 10, 50, 100, 200), which also can be termed as state dimension (s.dim). There are several observations. First, on the two private datasets (e.g., zx.math, zx.physics) which contain question texts, AKT-tx-tr performs better than DKT on two measurements, especially on 50 dimension knowledge state. Second, to further validate the effectiveness of the slipping and guessing modeling parts, we repeat the same experiments on the other seven public KT datasets described in section~\ref{datasets}. Obviously, we can find that although the performance of AKT-tx-tr is slightly lower (e.g., 0.4\% on AUC, 0.5\% on F1) than DKT on ASSISTments2015 and Synthetic, it gets better performance than DKT on five public KT datasets, especially on Static2011, it surpasses DKT by a large margin (e.g., $3.2\%$) on AUC metric. Third, there is an interesting discovery that the AUC of both DKT and AKT-tx-tr on dataset ASSISTments2017 is more than 0.7, but both of them have bad F1-score (e.g., no more than 0.5), that is DKT and AKT-tx-tr does not perform well, but a suitable threshold can be selected to obtain great AUC. Therefore, whether AUC is a reasonable measurement to evaluate knowledge tracing models is worth exploiting. In general, all the evidences observed above indicate that our framework can modeling slipping and guessing factors well for boosting knowledge tracing performance.

\begin{figure}[t]
	\centering
	\includegraphics[scale=0.63]{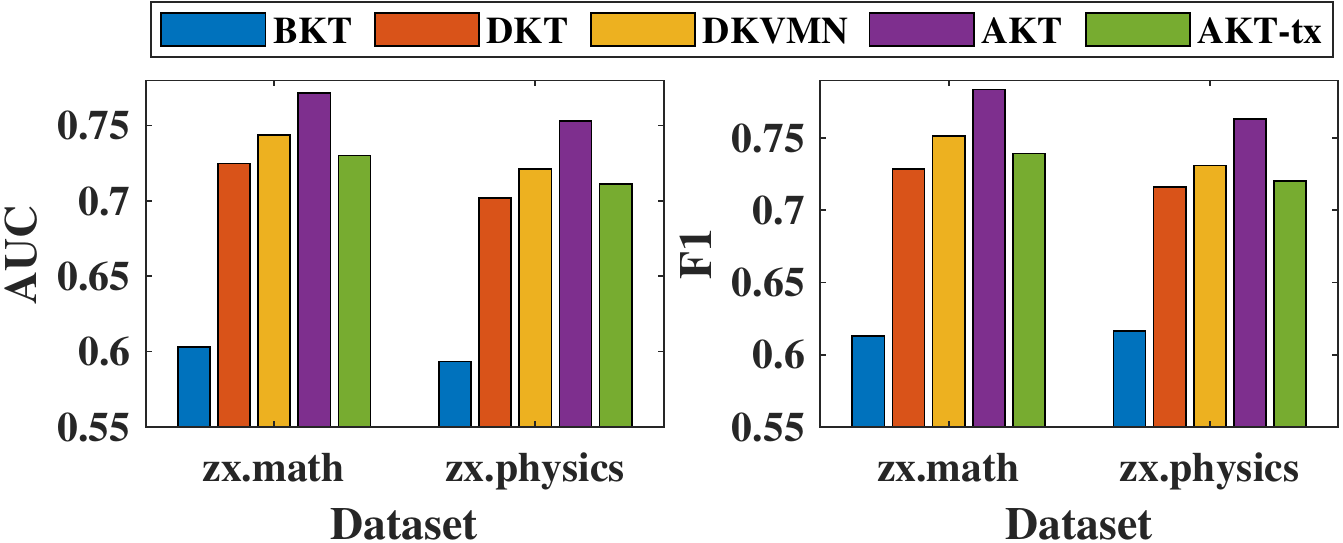}
	\caption{Question texts effectiveness validation.}
	\label{fig:texts}
\end{figure}

\subsubsection{Effectiveness of Question Texts}
To demonstrate the effectiveness of our framework of understanding the semantic of question texts for better question representation than one-hot to improve the KT performance. We conduct extensive experiments on AKT-tx  where "-" is a minus sign too with two private datasets (e.g., zx.math, zx.physics) since public datasets contain no question texts. The statistic information of the question texts of the two datasets are shown in Figure~\ref{fig:distribution}. For comparison, we adopt following three approaches as baselines:
\begin{itemize}
	\item \textbf{BKT}~\cite{corbett1994knowledge} is a Bayesian inference-based approach, in which the knowledge state is represented as a binary variable, it indicates the mastering of a single concept.
	\item \textbf{DKT}~\cite{Piech2015DeepKT} is an important KT model, which applies a recurrent neural network to model the student learning process for estimating students' mastery of concepts.
	\item \textbf{DKVMN}~\cite{Zhang2017Dynamic} which utilizes a key-value memory to extend the memory-augmented neural network is known as the state-of-the-art approach for KT.
\end{itemize}

The comparison results on each dataset on five different models including AKT are shown in Figure~\ref{fig:texts}. It is obvious that our adaptable method is able to boost the performance on each dataset comparing with baselines. Moreover, there are some observations. First, without understanding the question texts, the results of the variant AKT-tx (has no texts) is similar with DKT, that since the backbone of AKT-tx is the same as DKT except the slipping and guessing factors modeling parts. Second, AKT surpasses these comparison model by a large margin (e.g., 3.2\% on AUC, 3.24\% on F1), that because the rich question texts help train more parameters than one-hot representation. In general, the total observations prove that AKT can capture the semantic understanding of question texts effectively for boosting knowledge tracing performance.

In summary, our framework is helpful for the boosting knowledge tracing performance by modeling educational characteristics (e.g., slipping, guessing, question texts) for solving DAKT problem.

\begin{figure}[t]
	\centering
	\includegraphics[scale=0.67]{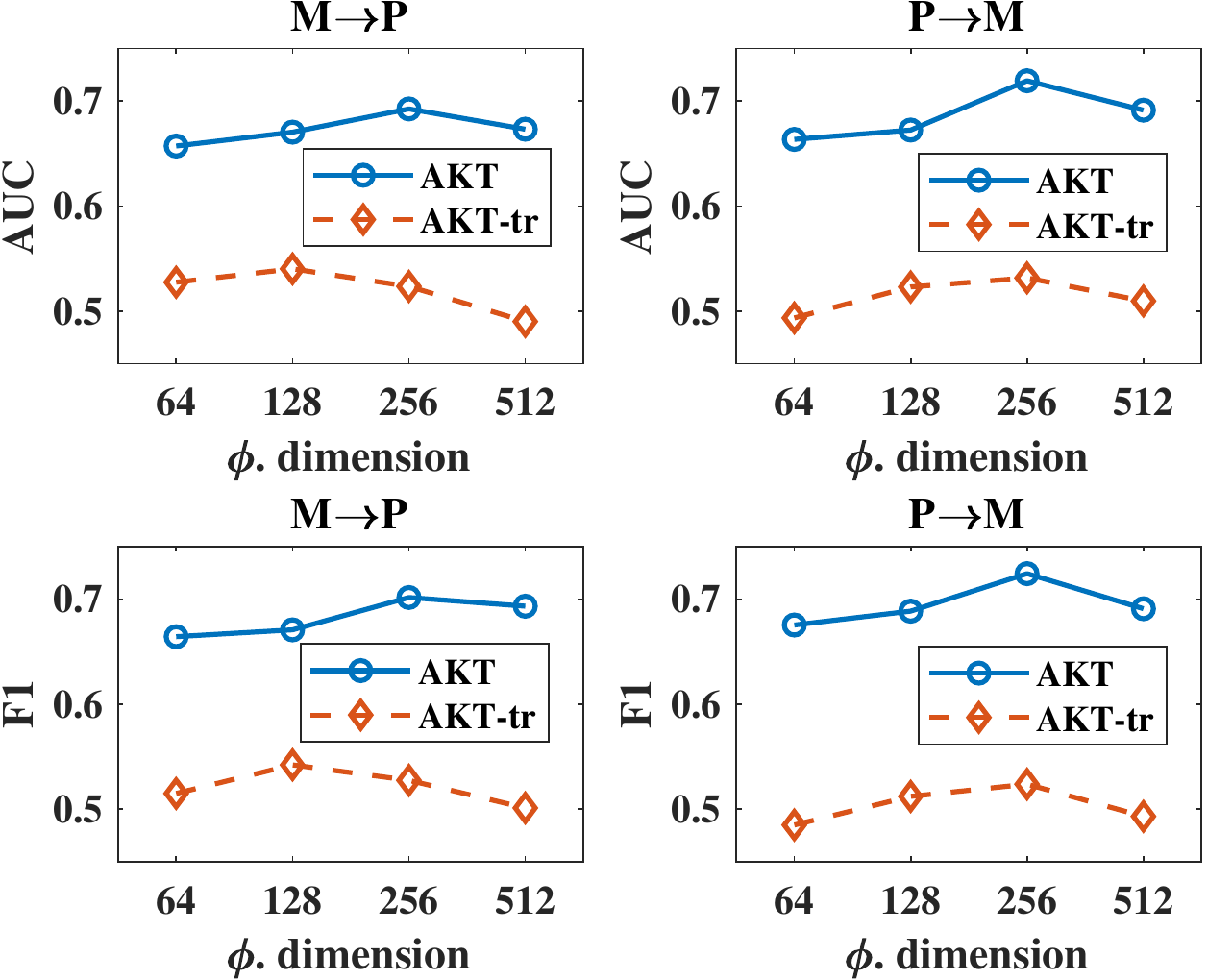}
	\caption{Evaluation of AKT for DAKT problem.}
	\label{fig:transfer}
\end{figure}

\subsection{Transferring Results}
In this section, we conduct extensive transferring experiments to evaluate the effectiveness of AKT for DAKT problem with subjects and schools as different domains. Moreover, there are two hyper-parameters in AKT that may impact its performance, therefore, we conduct sensitivity experiments to learn the effects of them. In transferring experiments, for each target domain, we randomly select a small amount of data as the limited labeled part $\mathcal{I}_{Tl}$, and the rest of it are used for validating.

\begin{table}[t]
	\centering
	\caption{The statistics of the selected schools.}
	\begin{tabular}{l|llll}  
		\toprule\toprule
		\multirow{1}{*}{Schools} & \multirow{1}{*}{A} & \multirow{1}{*}{B} & \multirow{1}{*}{C} & \multirow{1}{*}{D}\\
		\midrule\midrule
		\# Questions	& 708 	 	 & 694 	 	 & 611		& 764	\\
		\# Students		& 66 	 	 & 58 	 	 & 62		& 51	\\
		\# Interactions	& 4,673 	 & 4,066	 & 3,892	& 4,398	\\
		\bottomrule\bottomrule
	\end{tabular}
	\label{tab:schools}
\end{table}

\subsubsection{Subjects Transferring}
To validate that our AKT framework can achieve domain adaptation from source domain to target domain perfectly for DAKT problem, we adopt two subjects (e.g., math, physics) as two different domains by using datasets zx.math and zx.physics shown in Table~\ref{tab:statistics} and term them as M and P respectively. Therefore, there are $P_2^2=2$ transferring tasks $M\rightarrow P$ and $P\rightarrow M$ that need to be conducted, where $\rightarrow$ means transferring from the left to the right. Since there no existing methods for DAKT problem, we compare AKT with its variant AKT-tr where "-tr" means training AKT on source domain and applied to target domain directly without doing any transferring process. Moreover, to determine what dimension our learned adaptation layer should have, we try different widths from 64 to 512, stepping by a power of two each time. The experimental results are shown in Figure~\ref{fig:transfer}, we can find that both AUC and F1 of AKT-tr are approximate 0.5, these performances of it just like random prediction, that because no information is transferred from source domain to target domain, even causing negative transfer~\cite{pan2009survey}. However, both AUC and F1 of the adaptation method AKT are good on $M\rightarrow P$ and $P\rightarrow M$ tasks, especially, AKT performs best with 256 dimensions of adaptation layer. Totally, AKT is successful for addressing DAKT problem between subjects.

\begin{table*}[t]
	\centering
	\caption{The AUC of transferring information between four different schools which are sign as A, B, C and D.}
	\label{table:transfer_AUC}
	\begin{tabular}{cc|cccccccccccc}
		\toprule\toprule
		\multirow{1}{*}{Model}&\multirow{1}{*}{$\phi$.dim} & \multirow{1}{*}{A $\rightarrow$ B} & \multirow{1}{*}{B$\rightarrow$A} & \multirow{1}{*}{A$\rightarrow$C} & \multirow{1}{*}{C$\rightarrow$A} &\multirow{1}{*}{A$\rightarrow$D} & \multirow{1}{*}{D$\rightarrow$A} & \multirow{1}{*}{B$\rightarrow$C} & \multirow{1}{*}{C$\rightarrow$B} & \multirow{1}{*}{B$\rightarrow$D} & \multirow{1}{*}{D$\rightarrow$B} & \multirow{1}{*}{C$\rightarrow$D} & \multirow{1}{*}{D$\rightarrow$C}\\
		\midrule\midrule
		\multirow{4}{*}{AKT-tr}
		& 64  &      0.6576 &      0.6497 &      0.6482 &      0.6394 &      0.6511 &      0.6368 &      0.6401 &      0.6438 &      0.6381 &      0.6402 &      0.6432 &      0.6477\\
		& 128 &      0.6704 &      0.6637 &      0.6599 &      0.6413 &      0.6549 &      0.6537 &      0.6526 &      0.6518 &      0.6402 &      0.6507 &      0.6501 &      0.6588\\
		& 256 &      0.6924 &      0.6831 &      0.6816 &      0.6682 &      0.6670 &      0.6595 &      0.6645 &      0.6629 &      0.6523 &      0.6572 &      0.6457 &      0.6526\\
		& 512 &      0.6816 &      0.6715 &      0.6689 &      0.6527 &      0.6602 &      0.6498 &      0.6489 &      0.6507 &      0.6356 &      0.6474 &      0.6409 &      0.6501\\
		\cline{1-14}
		\multirow{4}{*}{AKT}
		& 64  &      0.7169 &      0.7086 &      0.7185 &      0.7073 &      0.7115 &      0.7009 &      0.6989 &      0.7070 &      0.7101 &      0.7178 &      0.7016 &      0.7151\\
		& 128 &      0.7232 &      0.7187 &      0.7224 & \bf{0.7234} & \bf{0.7247} &      0.7149 &      0.7120 & \bf{0.7236} &      0.7215 &      0.7286 & \bf{0.7203} & \bf{0.7250}\\
		& 256 & \bf{0.7338} & \bf{0.7242} & \bf{0.7305} &      0.7212 &      0.7160 & \bf{0.7205} & \bf{0.7208} &      0.7117 & \bf{0.7232} & \bf{0.7351} &      0.7162 &      0.7208\\
		& 512 &      0.7291 &      0.7163 &      0.7269 &      0.7036 &      0.7093 &      0.7154 &      0.7092 &      0.7106 &      0.7197 &      0.7263 &      0.7172 &      0.7113\\
		\bottomrule\bottomrule
	\end{tabular}
\end{table*}

\begin{table*}[t]
	\renewcommand\arraystretch{1.0}
	\centering
	\caption{The F1-score of transferring information between four different schools which are sign as A, B, C and D.}
	\label{table:transfer_F1}
	\begin{tabular}{cc|cccccccccccc}
		\toprule\toprule
		\multirow{1}{*}{Model}&\multirow{1}{*}{$\phi$.dim} & \multirow{1}{*}{A $\rightarrow$ B} & \multirow{1}{*}{B$\rightarrow$A} & \multirow{1}{*}{A$\rightarrow$C} & \multirow{1}{*}{C$\rightarrow$A} &\multirow{1}{*}{A$\rightarrow$D} & \multirow{1}{*}{D$\rightarrow$A} & \multirow{1}{*}{B$\rightarrow$C} & \multirow{1}{*}{C$\rightarrow$B} & \multirow{1}{*}{B$\rightarrow$D} & \multirow{1}{*}{D$\rightarrow$B} & \multirow{1}{*}{C$\rightarrow$D} & \multirow{1}{*}{D$\rightarrow$C}\\
		\midrule\midrule
		\multirow{4}{*}{AKT-tr}
		& 64  &      0.6721 &      0.6619 &      0.6541 &      0.6412 &      0.6640 &      0.6513 &      0.6384 &      0.6392 &      0.6358 &      0.6469 &      0.6509 &      0.6538\\
		& 128 &      0.6795 &      0.6726 &      0.6537 &      0.6434 &      0.6704 &      0.6539 &      0.6515 &      0.6491 &      0.6369 &      0.6491 &      0.6597 &      0.6613\\
		& 256 &      0.7095 &      0.6973 &      0.6792 &      0.6545 &      0.6743 &      0.6630 &      0.6563 &      0.6558 &      0.6495 &      0.6535 &      0.6513 &      0.6579\\
		& 512 &      0.6919 &      0.6836 &      0.6638 &      0.6421 &      0.6717 &      0.6507 &      0.6493 &      0.6369 &      0.6355 &      0.6484 &      0.6487 &      0.6566\\
		\cline{1-14}
		\multirow{4}{*}{AKT}
		& 64  &      0.7301 &      0.7206 &      0.7152 &      0.7128 &      0.7119 &      0.6940 &      0.7021 &      0.7091 &      0.6936 &      0.7020 &      0.6992 &      0.7019\\
		& 128 &      0.7338 &      0.7280 &      0.7214 & \bf{0.7244} &      0.7103 &      0.7051 &      0.7158 &      0.7109 &      0.7039 &      0.7133 & \bf{0.7125} & \bf{0.7171}\\
		& 256 & \bf{0.7385} & \bf{0.7319} & \bf{0.7251} &      0.7198 & \bf{0.7129} & \bf{0.7181} & \bf{0.7183} & \bf{0.7203} & \bf{0.7146} & \bf{0.7219} &      0.7074 &      0.7094\\
		& 512 &      0.7367 &      0.7213 &      0.7208 &      0.7091 &      0.7001 &      0.7103 &      0.7054 &      0.7098 &      0.6926 &      0.6974 &      0.7113 &      0.7025\\
		\bottomrule\bottomrule
	\end{tabular}
\end{table*}

\subsubsection{Schools Transferring}
To demonstrate the applicability of AKT for DAKT problem more completely. We further adopt four schools only selected from zx.math since zx.physics contains no information of school, and term them as A, B, C and D, the statistic of them are shown in Table~\ref{tab:schools}. Therefore, there are $P_4^2=12$ transferring tasks between four schools that need to be conducted. The same as subjects transferring, we term $\rightarrow$ as transferring from the left to the right, adopt AKT-tr as the comparison baseline and validate the effects of different adaptation layer dimension. The experimental results are shown in Table~\ref{table:transfer_AUC} and Table~\ref{table:transfer_F1}, they are AUC and F1-score of the models for the 12 transferring tasks. We can find that the performance of AKT-tr is better than random prediction (e.g., 0.5), that because the questions in the four schools are all selected from the same subject dataset zx.math, they are similar to some extent. Second, it is obvious that AKT obtains better transferring results than AKT-tr on both AUC and F1-score, that proves the effectiveness of AKT for solving DAKT problems between schools.

\begin{figure}[t]
	\centering
	\includegraphics[scale=0.67]{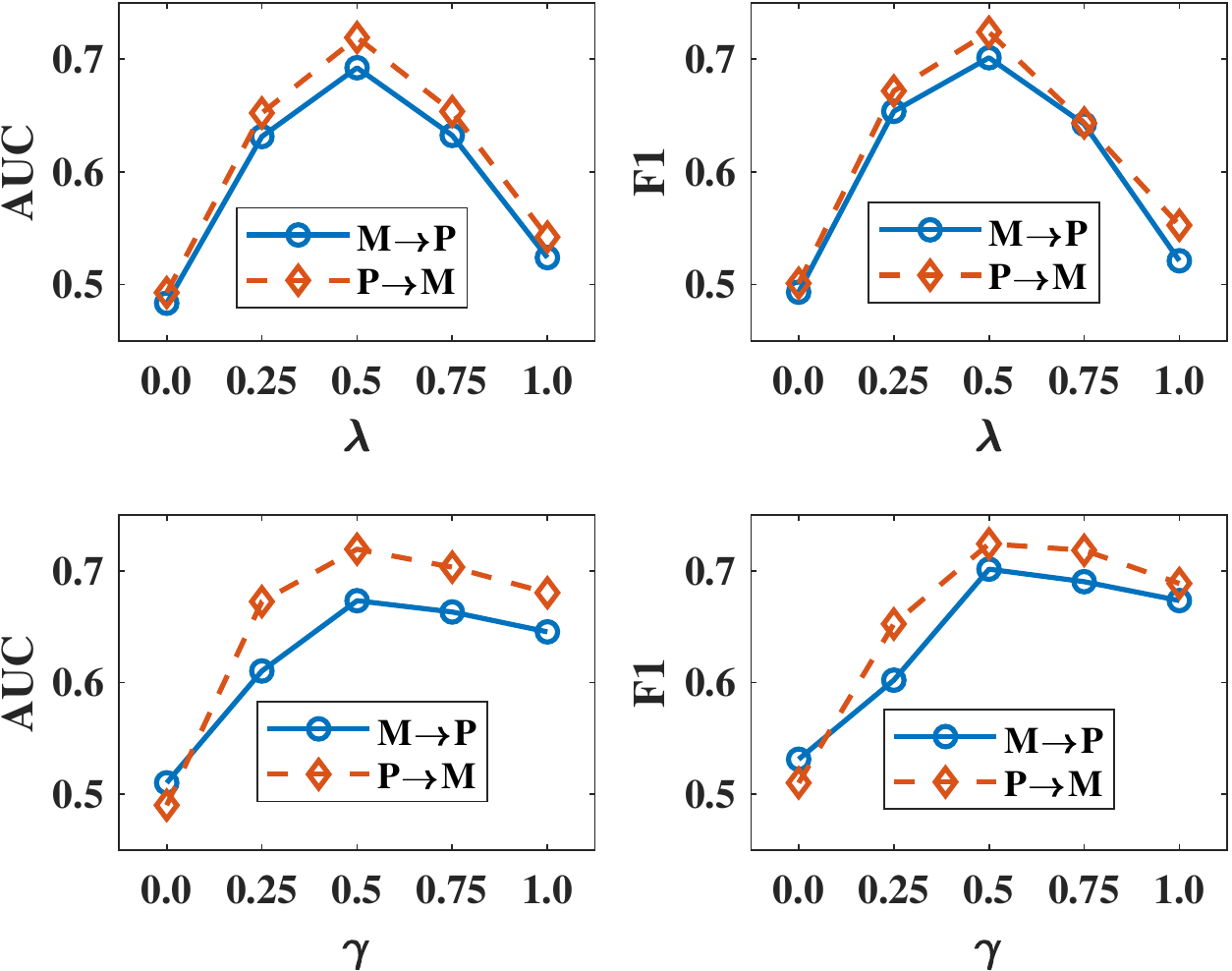}
	\caption{Sensitivity analysis of $\lambda$ and $\gamma$}
	\label{fig:lambda_gamma}
\end{figure}
\subsubsection{Sensitivity Analysis}
As presented in section~\ref{transfer}, there are two critical regularization factors $\lambda$ and $\gamma$, the well selected values of them can help the framework performs better, therefore, we conduct some empirical experiments to evaluate the sensitivity of them by setting different values (e.g., 0.0, 0.25, 0.5, 0.75, 1.0) of them on $M\rightarrow P$ and $P\rightarrow M$ tasks. The experimental results are shown in Figure~\ref{fig:lambda_gamma}. We can find that an appropriate $\lambda$ is of great significance for the performance of AKT, such as $\lambda=0.0$ means no source domain data are used, so the corresponding transferring results is not good, and the larger the $\lambda$ is, the more source domain data will be selected. The best value of $\lambda$ is around $0.5$ where both AUC and F1 of AKT are good. Meanwhile, it is obvious that $\gamma$ also plays an important role in AKT, since it balances the training loss and domain discrepancy. $\gamma=0.0$ means ignoring the domain discrepancy, that is the adaptation layer is abandoned rather than doing domain adaptation and the larger, the more important. Totally, the performance of AKT is sensitive to $\lambda$ and $\gamma$ to some extent, a good and patient selection of them will get ideal feedback.

\subsection{Discussion}
From the above experiments, it is obvious that AKT can perfectly address the DAKT problem. Moreover, we can find that the transferring results between different schools are better than the results between different subjects, which because the subjects for different schools are the same, the question texts of them are more similar. On the contrary, the question texts of different subjects are absolutely different. Therefore, transferring during different schools is easier than different subjects. Also, AKT can be applied not only between different subjects and schools but grades, even different subjects between different schools. However, there is still some room for improvement. First, AKT is sensitive to the hyper-parameters $\lambda$ and $\gamma$, how to model more reasonably, so that no need to manually set the $\lambda$ and $\gamma$ parameters is a problem worth studying. Second, the modeling process of slipping and guessing can be well designed in the future for more interpretable.

\section{Related Work}
Generally, our related works can be summarized as three categories, i.e. knowledge tracing, slipping and guessing factors of questions and domain adaptation techniques.

\subsection{Knowledge Tracing}
Conventional knowledge tracing models such as BKT~\cite{corbett1994knowledge} and PFA~\cite{pavlik2009performance} are the representative traditional approaches, they have been widely applied into many computer-aided education systems\cite{rowe2010modeling,schodde2017adaptive}. The BKT model has been proposed during 1990s. It is a two-stage dynamic Bayesian network in which student performance is the observed variable and student knowledge is the latent data, and it assumes a single knowledge is tested per question. Specifically, BKT adopts HMM to trace the knowledge states for each knowledge concept, and the knowledge state is represented by the hidden state, which is a binary variable that indicates whether or not the student masters the knowledge concept. Many variants of BKT have been proposed to solve these problems, such as~\cite{Hawkins2014LearningBK,Kser2017DynamicBN}. The PFA is a reconfiguration of LFA~\cite{Cen2006LearningFA} and offers higher sensitivity to student performance rather than student ability. It is proposed as an alternative to BKT, which relaxes the static knowledge assumption, models multiple knowledge concepts simultaneously with its basic structure and has the ability to handle the multiple knowledge concepts problems, but it cannot deal with the inherent dependency, e.g., "addition" is the prerequisite of "multiplication", and cannot provide a deep insight into student's latent knowledge state~\cite{Yeung2018AddressingTP}.

Since deep learning approaches outperform the traditional models in many areas (e.g., natural language processing), some researchers attempt to leverage deep learning techniques to predict knowledge state more precisely. 
The {\em deep knowledge tracing} (DKT) was the first attempt, to the best of our knowledge, to
utilize recurrent neural networks (e.g., RNN and LSTM) to model student’s learning process for predicting her performance~\cite{Piech2015DeepKT}, It outperforms conventional methods in most of the application scenes because of its nonlinear input-to-state and state-to-state transitions. Following DKT, there are amounts of studies propose the variant of it, such as DKT+forgetting~\cite{nagatani2019augmenting} aggregates extra question information (e.g., forgetting) and DKT-tree~\cite{yang2018implicit} introduces the learning time to DKT.

\subsection{Slipping and Guessing Factors}
For more precisely simulating student learning process, it is necessary to completely model the situation, i.e.slipping and guessing that students may encounter when answering questions, which are strongly interpretable for student knowledge state. There are two mediating parameters (e.g., slipping, guessing) in BKT~\cite{Baker2008MoreAS}, the slipping parameter acknowledges that even student who masters a knowledge concept can make an occasional mistake, and the guessing parameter represents the fact that a student may sometimes generate a correct response, in spite of not knowing the correct knowledge concept. The {\em item response theory} (IRT) also adopts a parameter $c$ to represent the guessing factors of items, which mirrors the probability that a student gets a right answer when his latent trait is low\cite{rasch1960studies}. Wu et al. introduces slipping and guessing factors into {\em cognitive diagnosis model} (CDM) to compute examinees' problem mastery and generates examinees' observable scores on problems~\cite{Wu2015CognitiveMF}, and propose the gaming factor which can be seen as guessing factors for student learning modeling, to obtain more precise and reasonable cognitive analysis\cite{Wu2017KnowledgeOG}. Totally, there are only several studies that exploit the slipping and guessing factors of the question, and almost have not been studied in deep knowledge tracing area.

\subsection{Domain Adaptation Techniques}
Domain adaptation is a branch of transferring learning, which focuses on dealing with the problem that the feature and label space are consistent respectively while the feature distribution is inconsistent. It has been widely applied in amounts of directions such as text classification~\cite{Lu:2014:SFT:2893873.2893894} and object detection~\cite{Chen2018DomainAF}, etc. Up to now, there are many transfer learning techniques which can be categorized as instance based~\cite{tan2017distant}, feature based~\cite{long2017deep}, model based~\cite{long2016deep} and relationship based~\cite{davis2009deep} approaches. As one of the simplest transfer learning techniques, fine-tuning has been adopted to address some real problems, i.e. Daniel et al. introduce fine-tuning to train a deep learning model for optical coherence tomography images classification in cell journal~\cite{kermany2018identifying}. However, to the best of our knowledge, there is no attempt to propose the {\em domain adaptation for knowledge tracing} (DAKT) problem. Therefore, we design a general transferable framework for knowledge tracing, this is the first time to do some attempts in DAKT problem by leveraging transfer learning techniques in this study.

\section{Conclusion}
In this paper, we addressed the issues of existing KT methods and studied the {\em domain adaptation for knowledge tracing} (DAKT) problem. We proposed an adaptable knowledge tracing (AKT) framework which integrates slipping and guessing factors into framework for more reasonable knowledge state modeling results, leverages the semantic understanding of question texts to replace the one-hot representation for more precise knowledge tracing, and contains {\em instance selection via pre-training}, {\em domain discrepancy minimizing} and {\em fine-tuning} three transferring steps for DAKT problem.
Extensive experimental results demonstrated that AKT was successful for DAKT problem and superior to comparison methods for better KT results. We hope this study builds a solid baseline for DAKT problem to promote more researches in this field.



\bibliographystyle{ACM-Reference-Format}
\bibliography{www2020}

\end{document}